\documentclass[10pt]{article} 
\usepackage[preprint]{tmlr}

\usepackage{amsmath,amsfonts,bm}

\def\eqref#1{equation~\ref{#1}}

\def\1{\bm{1}}

\DeclareMathAlphabet{\mathsfit}{\encodingdefault}{\sfdefault}{m}{sl}
\SetMathAlphabet{\mathsfit}{bold}{\encodingdefault}{\sfdefault}{bx}{n}

\usepackage{microtype}
\usepackage{graphicx}
\usepackage{subcaption}
\usepackage{booktabs} 
\usepackage{mathtools}         
  \mathtoolsset{showonlyrefs}
\usepackage{hyperref}
\usepackage{hyperref}
\usepackage{url}

\newcommand{\Lcal}{\mathcal{L}}
\newcommand{\Rcal}{\mathcal{R}}
\newcommand{\Rspace}{\mathbb{R}}
\newcommand{\Thetab}{\boldsymbol{\Theta}}

\usepackage{amsmath}
\usepackage{amssymb}
\usepackage{mathtools}
\usepackage{amsthm}

\usepackage[capitalize,noabbrev]{cleveref}

\theoremstyle{plain}
\newtheorem{theorem}{Theorem}[section]

\newtheorem{lemma}[theorem]{Lemma}

\theoremstyle{definition}
\newtheorem{definition}[theorem]{Definition}
\newtheorem{assumption}[theorem]{Assumption}
\theoremstyle{remark}
\newtheorem{remark}[theorem]{Remark}

\title{From Sublinear to Linear: Local Convergence in Finite-Width Networks via Locally Polyak-Lojasiewicz Regions}

\author{\name Agnideep Aich \email agnideep@stanford.edu \\
      \addr Department of Emergency Medicine\\
      Stanford University
      \AND
      \name Ashit Baran Aich \email aichnsou@gmail.com \\
      \addr Department of Statistics\\
      Formerly of Presidency College\\
      Kolkata, India
      \AND
      \name Bruce A. Wade \email bruce.wade@louisiana.edu \\
      \addr Department of Mathematics\\
      University of Louisiana at Lafayette}

\def\month{MM}  
\def\year{YYYY} 
\def\openreview{\url{https://openreview.net/forum?id=XXXX}} 

\begin{document}

\maketitle

\begin{abstract}
We study local linear convergence of gradient descent for finite-width feedforward networks under the squared empirical loss. Prior work shows that GD can remain confined to a Locally Quasi-Convex Region (LQCR) around initialization, but only gives a sublinear rate. We show that if the empirical Neural Tangent Kernel is positive at initialization, Lipschitz stable on the LQCR, and compatible with the LQCR radius, then the squared loss satisfies a local Polyak-{\L}ojasiewicz inequality with constant $\mu = \lambda_0 - L_\Theta r(\Rcal) > 0$. Combined with fixed-step iterate containment in the LQCR, imposed as a hypothesis in the linear-rate theorem, this yields linear convergence on the region. The LQCR supplies localization; fixed-step containment is imposed as a hypothesis in the linear-rate theorem; and the PL inequality comes from NTK conditioning under squared loss. The result is therefore a sufficient local condition, not a claim that this mechanism is necessary or unique for fast convergence. Empirically, we probe the theory through NTK spectral gap, parameter drift, empirical PL ratio, and suboptimality decay. On binary MNIST, the NTK remains positive, the PL ratio has a positive lower envelope, and the loss shows geometric decay on the stable regime. In a width ablation, the fixed-step width-$1024$ run leaves the local regime; reducing the step size lowers final drift from $1.870$ to $0.158$, restores the observed local-regime diagnostics, and yields the largest empirical PL-ratio lower envelope observed in the study. A CNN robustness check on a CIFAR-10 subset shows the PL-ratio envelope remains positive across three seeds, with a positive lower envelope across all three seeds on the stable regime.
\end{abstract}

\section{Introduction}
\label{sec:intro}

Gradient descent on deep neural network loss landscapes is non-convex, yet often converges far faster in practice than classical worst-case guarantees suggest. A long line of theoretical work has narrowed the gap between this empirical phenomenon and provable guarantees, but typically by either restricting to specific architectures, requiring extreme width, or assuming favorable structure such as a Polyak-{\L}ojasiewicz (PL) \citep{Polyak1963} condition directly. Among the analyses that operate in finite-width (taken to mean without an asymptotic procedure), standard-architecture settings, the Locally Quasi-Convex Region (LQCR) framework of \citep{Aich2025} establishes that a neighborhood around standard initializations exists in which GD is provably confined under the step-size assumptions of that framework, and converges to a stationary point at a sublinear rate. The LQCR construction is local, geometric, and applies to standard feedforward networks, but the rate it guarantees inside the region remains sublinear; the gap between this rate and the near-exponential decay observed in our experiments is the starting point of this article.

\textbf{The question.} If GD provably stays inside an LQCR with a known radius and a sublinear rate, is there a natural \emph{additional} local condition under which the rate inside that same region improves to linear? Such a condition should enter the analysis explicitly rather than merely be assumed as a global property, and it should apply to the local region the LQCR already characterizes rather than to the whole parameter space.

\textbf{The answer, and what it is not.} We show that the squared empirical loss satisfies a Polyak-{\L}ojasiewicz inequality on the LQCR whenever the empirical Neural Tangent Kernel \citep{Jacot2018} (i) is pointwise positive at initialization and (ii) is Lipschitz continuous on the region, with a compatibility condition tying the region radius to the initial spectral gap. The mechanism is a one-line squared-loss identity: $\|\nabla\Lcal(\theta)\|^2 = \tfrac{1}{n}(f_\theta - y)^\top \Thetab_\theta (f_\theta - y)$, combined with a uniform spectral lower bound on $\Thetab_\theta$ across the region derived via Weyl's inequality. The LQCR enters the result through its radius $r(\Rcal)$, which controls how far NTK drift can erode the initial spectral gap, and through the local containment regime identified by the LQCR framework. In the fixed-step theorem of this paper, iterate containment is imposed as a hypothesis; under that containment hypothesis, GD converges linearly on the LQCR. We call such a region a \emph{Locally Polyak--{\L}ojasiewicz Region} (LPLR).

This is a sufficient-condition result, scoped to the squared loss and to the local regime in which the iterates remain inside an LQCR satisfying the NTK conditioning assumption. It is not a claim that the LPLR mechanism is necessary for fast convergence in deep networks, nor that it is the only route by which linear-rate behavior can arise in practice. We position the result as a local, finite-width sufficient condition rather than a universal explanation. Linear-convergence mechanisms based on NTK, Gram-matrix, or Jacobian conditioning are well established in overparameterized neural-network theory \citep{Du2019, AllenZhu2019, Oymak2019}; our contribution is to derive the same PL mechanism locally, inside a geometrically characterized finite-width region, from a pointwise spectral condition at initialization plus Lipschitz stability.

\textbf{Empirical strategy.} Beyond the theory, we argue for a different empirical methodology than is typical in this literature. The latent variables of the framework, namely the NTK spectral gap, the parameter drift, the empirical PL ratio, and the iterate-region condition, are all directly measurable along a training trajectory. Rather than fitting log-log slopes to loss curves and reporting that they look polynomial, we track each latent variable in turn and ask whether it behaves as the theory predicts. This style of validation is more demanding than reporting consequences alone: it can disconfirm the framework if any latent variable behaves contrary to the theory. We adopt it precisely because reviewers and readers should be able to see to what the theory commits, on the runs where its assumptions are intended to apply.

\textbf{Contributions.}
\begin{enumerate}
    \item \textbf{A local PL inequality from local NTK conditioning, with explicit decomposition of roles.} We prove that on an LQCR around initialization, pointwise positivity of the empirical NTK at $\theta_0$ together with Lipschitz continuity across the region yields a PL inequality for the squared loss, with PL constant $\mu = \lambda_0 - L_\Theta r(\Rcal) > 0$ (Theorem~\ref{thm:lplr_existence}). The role of each ingredient is explicit: pointwise positivity is the initial spectral budget, Lipschitz continuity sets the rate at which NTK drift can erode it, and the LQCR radius $r(\Rcal)$ supplies the maximum distance over which propagation is needed. Linear convergence of GD on the LPLR then follows by combining the PL inequality with fixed-step iterate containment in the region, imposed as a hypothesis in the linear-rate theorem, under the step-size requirement $\eta \le 1/L$ (Theorem~\ref{thm:linear_conv}).

    \item \textbf{Localization through the LQCR, not derivation from it.} We position the LQCR explicitly as a containment-and-localization device rather than as the source of the PL inequality. The PL inequality is a consequence of NTK conditioning under squared loss, not of the LQCR curvature or descent conditions. We emphasize this decomposition both to be accurate about what each ingredient contributes and to avoid the overclaim that local geometry alone produces fast rates.

    \item \textbf{Sufficient, not necessary, and not universal.} We would like to state that the LPLR is a sufficient local condition for linear convergence on the squared loss, and not a claim about the dominant or unique mechanism for fast deep-network optimization. We do not claim the LPLR mechanism explains fast convergence in practice in general; we identify a specific local regime in which it provably suffices.

    \item \textbf{An empirical methodology that probes latent variables, not just consequences.} We design experiments around the actual latent variables of the theory (NTK spectral gap, parameter drift, empirical PL ratio, iterate containment, loss decay), and validate the framework on a controlled MLP setting, on a width ablation that maps the boundary of the local regime and demonstrates its recovery under an adapted step size, and on a CNN robustness check under mini-batch SGD with cosine annealing.

    \item \textbf{A sharp empirical demonstration of where the local regime breaks and recovers.} The width ablation includes a configuration at $(m, \eta) = (1024, 10^{-3})$ where the iterates leave the lazy regime, parameter drift jumps by an order of magnitude, NTK conditioning is substantially eroded, and the empirical PL constant degrades, which is exactly the kind of failure mode expected when the containment hypothesis of Theorem~\ref{thm:linear_conv} is not maintained. Reducing the step size to $\eta = 5 \times 10^{-4}$ restores the observed local-regime diagnostics and yields the largest empirical PL-ratio lower envelope of the study. We view this paired result as one of the more informative outcomes of the paper: the framework's predictions are confirmed in the regime it applies to, and its failure mode is itself a confirmation of the step-size condition the theorem requires.
\end{enumerate}

\textbf{Paper organization.} Section~\ref{sec:related} positions the contribution against the NTK, LQCR, and PL literature. Section~\ref{sec:notation} fixes notation. Section~\ref{sec:framework} sets up the LPLR concept and recalls the LQCR ingredients used later. Section~\ref{sec:results} proves the local PL inequality (Theorem~\ref{thm:lplr_existence}) and the linear-rate guarantee (Theorem~\ref{thm:linear_conv}), and states the scope of the result. Section~\ref{sec:experiments} presents the empirical validation across the MLP, width ablation, and CNN settings. Section~\ref{sec:conclusion} provides the discussion and conclusion.
\section{Related Work}
\label{sec:related}

We position the contribution against three adjacent literatures: PL-type conditions in deep learning theory, Neural Tangent Kernel analyses of finite-width networks, and local landscape characterizations including the LQCR framework. In each case we identify what is shared with our setting and what is distinct, with particular care to avoid overstating novelty relative to existing NTK-style linear-convergence results.

\subsection{PL-type conditions in deep learning theory}

The Polyak-{\L}ojasiewicz condition \citep{Polyak1963} provides one of the cleanest routes from a non-convex optimization problem to a linear convergence rate. It has also been used as a common condition for analyzing gradient, proximal-gradient, coordinate, stochastic, and variance-reduced first-order methods, while related relaxed strong-convexity and error-bound conditions have been used to prove linear convergence in non-strongly convex settings \citep{Karimi2016, Necoara2015}. Recent work has applied PL-style or {\L}ojasiewicz-type conditions to deep learning in three broadly distinct ways.

First, \emph{architectural} approaches design networks that satisfy favorable optimization inequalities by construction. \citep{wang2024monotone} introduce BiLipNet, a bi-Lipschitz invertible architecture, and PLNet, a scalar-output network constructed from a BiLipNet and a quadratic potential; they show that PLNet satisfies the Polyak-{\L}ojasiewicz condition. These results give clean guarantees, but the guarantees are tied to the proposed architecture rather than to standard feedforward networks.

Second, \emph{assumption-based} approaches posit a PL or {\L}ojasiewicz-type condition as a hypothesis and derive consequences for stochastic or deterministic optimization. \citep{an2024convergence} prove local convergence of SGD with positive probability under a local {\L}ojasiewicz condition and an additional local structural assumption on the loss landscape. \citep{delarue2025genericity} study genericity of local PL inequalities for entropic mean-field neural ODEs. These works yield strong conclusions when their hypotheses hold, but they are not direct sufficient-condition results for the standard finite-width nonlinear feedforward setting considered here.

Third, \emph{derivation-based} approaches identify concrete settings in which PL or {\L}ojasiewicz-type behavior follows from underlying structure. \citep{xu2025local} establish local PL and descent inequalities for overparameterized two-layer linear networks under relaxed assumptions on step size, width, and initialization. \citep{zhou2021local} develop a local convergence theory for mildly over-parameterized two-layer neural networks, showing convergence in a local regime once the loss is below a suitable threshold. These works are therefore closest in spirit to ours, but remain restricted to two-layer or linear settings.

Our contribution sits in this third category, but for finite-width nonlinear feedforward networks: we identify a concrete local condition on the empirical NTK under which the squared loss satisfies a PL inequality on the LQCR.

\subsection{Neural Tangent Kernel and linear convergence}

The Neural Tangent Kernel \citep{Jacot2018} provides a central bridge between training dynamics, kernel conditioning, and convergence for wide neural networks. In the infinite-width limit, \citep{Jacot2018} show that the network function evolves according to kernel gradient descent with respect to the NTK, and that convergence can be related to positive-definiteness of the limiting NTK. For squared loss, this connection is especially direct: lower bounds on the smallest eigenvalue of the empirical tangent kernel control the residual through the squared-gradient norm, which is the mechanism used in our local PL argument.

This mechanism is closely related to several overparameterization results. \citep{Du2019} prove that randomly initialized GD converges linearly to a global optimum for overparameterized two-layer ReLU networks, with the rate controlled by the least eigenvalue of a Gram matrix. \citep{Du2018} prove linear convergence to zero training loss for deep overparameterized networks with residual connections, using stability of the architecture-induced Gram matrix. \citep{AllenZhu2019} extend the analysis to multi-layer networks under polynomial width requirements and derive polynomial-time convergence to global minima. \citep{Arora2019a} study exact computation of infinite-width NTKs, including convolutional NTKs, and give a non-asymptotic result connecting sufficiently wide trained networks to NTK kernel regression. \citep{Lee2019} show that wide neural networks of any depth evolve as linearized models under gradient descent. \citep{Liu2020} develop a PL$^*$ framework for overparameterized nonlinear systems and relate the condition to the spectrum of the tangent kernel. \citep{Oymak2019} give geometric-rate convergence guarantees for moderately overparameterized shallow networks. \citep{novak2022fast} address efficient computation of finite-width NTKs.

The shared theme in the optimization results above is that suitable Gram-matrix, tangent-kernel, or Jacobian conditioning can yield fast loss decay in overparameterized or infinite-width regimes. Our contribution is not the discovery of this mechanism, but its localization to a finite-width region with explicit radius. Instead of requiring a global or infinite-width conditioning argument, we assume a pointwise spectral condition at initialization, propagate it across the LQCR via Lipschitz continuity and Weyl's inequality, and obtain a PL inequality on that region. The width requirement driving the LQCR construction of \citep{Aich2025} is $\Omega(L^3)$ rather than polynomial in $n$, so the LPLR result inherits this local finite-width regime. We view the relationship to prior NTK-style linear-convergence results as one of localization and scope refinement rather than mechanism replacement.

\subsection{Local landscape geometry and the LQCR framework}

Landscape-geometry analyses of deep networks ask when the loss surface admits structural properties that help explain optimization success. Early results related neural-network loss surfaces to spin-glass models under simplifying assumptions, showing that poor local minima become exponentially rare in the corresponding large-size decoupled model \citep{choromanska2015loss}. For deep linear networks, \citep{kawaguchi2016deep} prove that every local minimum is global. For nonlinear networks, \citep{nguyen2017loss} show that, under a wide-layer condition and additional architectural assumptions, almost all local minima are globally optimal. \citep{petzka2021nonattracting} show that suboptimal local minima can exist even in deep and wide networks, but may lie in non-attracting regions connected to lower-loss paths. \citep{kawaguchi2020practical} prove trainability guarantees for deep nonlinear networks with a number of parameters growing essentially linearly in the number of samples, together with generalization guarantees for natural datasets in their setting.

The Locally Quasi-Convex Region framework of \citep{Aich2025} contributes a finite-width, local-geometry construction: under standard initialization and a hidden-width condition written informally as $m=\Omega(L^3)$, an LQCR of explicit radius $r(\Rcal)$ exists around the initialization point. In that framework, gradient descent with the prescribed decaying step size $\eta_t=\eta_0/(1+t)^{1/3}$ and $\eta_0\le 1/\alpha$ remains inside the region and satisfies a sublinear stationarity guarantee. Our work uses the LQCR not as a source of fast rates, since that would conflate the LQCR curvature and descent conditions with NTK conditioning, but as a localization and containment device on which an independent NTK-based PL argument can be run. The decomposition is deliberate: the LQCR supplies the region and, in its original decaying-step result, a containment mechanism; our fixed-step theorem assumes containment on that region; NTK conditioning under squared loss supplies the PL inequality; the rate is the consequence.

\subsection{Other mechanisms for PL-like behavior}

Beyond the NTK route, several alternative sources of PL-type behavior in neural networks deserve mention. \citep{wang2024monotone} provide a constructive route through architectural design by introducing PLNet, a scalar-output architecture that satisfies the Polyak-{\L}ojasiewicz condition by construction. Mean-field and feature-learning analyses operate in regimes that differ substantially from the lazy/NTK picture and may apply where ours does not. We view our framework as one specific local sufficient condition under which fast rates provably arise, and not as a general explanation of fast convergence in practice.
\section{Notation}
\label{sec:notation}

We introduce the notation used throughout the analysis. We consider a feedforward neural network
\[
f_\theta : \Rspace^d \to \Rspace
\]
with $D$ affine layers and layer widths $m_0,\ldots,m_D$, where $m_0=d$ is the input dimension and $m_D=1$ is the scalar output dimension. The parameter vector $\theta \in \Rspace^p$ collects all weights and biases. When the hidden layers have a common width, we write $m_1=\cdots=m_{D-1}=m$ and refer to $m$ as the hidden width. Given training data $\{(x_i,y_i)\}_{i=1}^n$ with $x_i \in \Rspace^d$, we study the squared empirical loss
\begin{align}
\Lcal(\theta)
=
\frac{1}{2n}\sum_{i=1}^n \big(f_\theta(x_i)-y_i\big)^2.
\end{align}
Gradient descent is written as
\begin{align}
\theta^{(t+1)}
=
\theta^{(t)}-\eta \nabla \Lcal(\theta^{(t)}),
\end{align}
where $\eta>0$ is the step size. We write $\theta_0$ for the initialization point and $\theta^{(t)}$ for the iterate at step $t$. The loss $\Lcal$ is assumed to be $L$-smooth on the local region considered below.

Our analysis takes place inside a local region $\Rcal \subset \Rspace^p$ around $\theta_0$. In the LQCR framework of \citep{Aich2025}, this region is parameterized by a curvature constant $\alpha$, a descent constant $\gamma$, and a radius $r(\Rcal)$. We denote the minimum loss value inside $\Rcal$ by
\begin{align}
\Lcal_\Rcal^*
=
\min_{\theta \in \Rcal} \Lcal(\theta),
\end{align}
and the global infimum by
\begin{align}
\Lcal^*
=
\inf_{\theta \in \Rspace^p} \Lcal(\theta).
\end{align}
In the experiments, $\Lcal_\Rcal^*$ is approximated by the terminal training loss
\begin{align}
\widehat{\Lcal}_\Rcal^*
:=
\Lcal(\theta^{(T)}).
\end{align}

Let $J_\theta$ denote the Jacobian of the network outputs on the training inputs with respect to the parameters. The empirical Neural Tangent Kernel matrix is
\begin{align}
\Thetab_\theta
=
\frac{1}{n}J_\theta J_\theta^\top
\in \Rspace^{n\times n}.
\end{align}
We denote its smallest eigenvalue by $\lambda_{\min}(\Thetab_\theta)$, and write
\begin{align}
\lambda_0
:=
\lambda_{\min}(\Thetab_{\theta_0})
\end{align}
for the value at initialization. The empirical NTK is assumed to be $L_\Theta$-Lipschitz on $\Rcal$, meaning that
\begin{align}
\|\Thetab_\theta-\Thetab_\phi\|
\le
L_\Theta\|\theta-\phi\|
\quad
\text{for all } \theta,\phi\in\Rcal.
\end{align}
Under the local NTK conditioning assumption introduced in Section~\ref{sec:results} and via Lemma~\ref{lem:weyl}, we will write
\begin{align}
\lambda_\Rcal
:=
\lambda_0 - L_\Theta r(\Rcal)
>
0.
\end{align}
This quantity becomes the local Polyak-{\L}ojasiewicz constant $\mu=\lambda_\Rcal$ in Theorem~\ref{thm:lplr_existence}.
\section{From Local Geometry to a Local Polyak-Lojasiewicz Condition}
\label{sec:framework}

This section sets up the two ingredients we need to obtain linear convergence of gradient descent in a finite-width setting: a local region where the iterates of GD are confined, and a local error-bound condition that controls suboptimality by the squared gradient norm. The first ingredient is supplied by the Locally Quasi-Convex Region (LQCR) framework of \citep{Aich2025}, which gives us a geometrically characterized region around initialization and, under its prescribed decaying step-size schedule, an iterate-confinement guarantee. The second ingredient, a local Polyak-Lojasiewicz (PL) inequality, will be obtained in Section~\ref{sec:results} from a local stability condition on the Neural Tangent Kernel. We emphasize at the outset that, in our framework, the LQCR is not the source of the PL condition. The PL condition will be derived from NTK conditioning under squared loss. The role of the LQCR is to define the local region in which that spectral condition is assumed to hold; in the fixed-step theorem below, remaining inside that region is imposed as a hypothesis.

The PL condition, introduced by Polyak \citep{Polyak1963}, gives linear convergence for gradient descent under suitable smoothness assumptions, and has since been used to analyze broader first-order methods \citep{Karimi2016}. In its global form,
\begin{align}
\frac{1}{2} ||\nabla \Lcal(\theta)||^2 \ge \mu (\Lcal(\theta) - \Lcal^*),
\label{eq:pl_global_restated}
\end{align}
where $\Lcal^*$ is the global minimum, the inequality is too strong to expect for general deep network objectives. What we will use is a localized version that only needs to hold inside the region the optimizer actually visits.

\begin{definition}[Locally Polyak-Lojasiewicz (LPL) Region]
\label{def:lplr_restated}
A region $\Rcal \subset \mathbb{R}^p$ is a $(\mu, \Rcal)$-Locally Polyak-Lojasiewicz Region (LPLR) if
\begin{align}
\frac{1}{2} ||\nabla \Lcal(\theta)||^2 \ge \mu (\Lcal(\theta) - \Lcal_{\Rcal}^*) \quad \forall \theta \in \Rcal,
\label{eq:pl_local_restated}
\end{align}
where $\Lcal_{\Rcal}^* = \min_{\theta \in \Rcal} \Lcal(\theta)$ is the minimum value of the loss within $\Rcal$.
\end{definition}

The LPLR is a property of the loss function on a region. It does not by itself say where such a region exists, how large it is, or whether GD stays inside it. Those questions are answered separately by the LQCR framework of \citep{Aich2025}, which we now recall.

\begin{definition}[Locally Quasi-Convex Region \citep{Aich2025}]
\label{def:lqcr}
A region $\Rcal \subset \mathbb{R}^p$ is an $(\alpha,\gamma)$-LQCR if for all $\theta, \phi \in \Rcal$:
\begin{enumerate}
    \item \textbf{Curvature Bound:} $\Lcal(\phi) \ge \Lcal(\theta) + \nabla \Lcal(\theta)^\top(\phi - \theta) - \frac{\alpha}{2}\|\phi - \theta\|^2$.
    \item \textbf{Descent Condition:} If $\Lcal(\phi) < \Lcal(\theta)$, then $\langle -\nabla \Lcal(\theta), \phi - \theta \rangle \ge \gamma \|\phi - \theta\|\, \|\nabla \Lcal(\theta)\|$.
\end{enumerate}
\end{definition}

Two facts from \citep{Aich2025} are what we will actually use later. First, under standard initialization schemes and hidden width $m = \Omega(L^3)$, an $(\alpha,\gamma)$-LQCR exists around initialization with an explicit radius $r(\Rcal)$ depending on depth, width, and activation bounds. Second, under the prescribed decaying step size
$\eta_t=\eta_0/(1+t)^{1/3}$ with $\eta_0\le 1/\alpha$,
the LQCR result guarantees, with probability at least $1-\delta$,
that the iterates remain inside this region. These two facts give us the local region and the containment guarantee. They do not, on their own, give us a PL inequality, and we do not claim they do. The PL inequality will come from a separate assumption on the Neural Tangent Kernel that we state in the next section.

\begin{remark}[Sign convention in the descent condition]
\label{rem:sign_convention}
The descent condition in Definition~\ref{def:lqcr} is stated using $-\nabla \Lcal(\theta)$ rather than $\nabla \Lcal(\theta)$. This is the geometrically correct form: when $\Lcal(\phi) < \Lcal(\theta)$, the steepest-descent direction $-\nabla \Lcal(\theta)$ should have positive projection along the displacement $(\phi - \theta)$ from $\theta$ toward the lower-loss point $\phi$. This convention corrects a sign in the original LQCR statement; the corrected version is consistent with the submitted corrigendum to \citep{Aich2025}. All subsequent results in this paper use this sign convention.
\end{remark}

The bridge from this geometric picture to the PL condition runs through the empirical Neural Tangent Kernel \citep{Jacot2018}. Recall that for a network with Jacobian $J_\theta(x)$, the empirical NTK is $\Thetab_\theta(x, x') = J_\theta(x) J_\theta(x')^\top$, and the corresponding $n \times n$ Gram matrix over the training inputs is $\Thetab_\theta = \tfrac{1}{n} J_\theta J_\theta^\top$. The bridge itself is a one-line identity for squared loss: when the smallest eigenvalue of $\Thetab_\theta$ is bounded below by some $\lambda > 0$, the squared gradient norm controls the residual, and the residual controls the suboptimality gap. We make both the identity and the conditions under which $\lambda_{\min}(\Thetab_\theta) > 0$ holds across the region precise in Section~\ref{sec:results}.

The picture to keep in mind for the rest of the paper is therefore the following. The LQCR supplies the region $\Rcal$ with an explicit radius $r(\Rcal)$ and, in its original decaying-step result, a containment mechanism. NTK conditioning, under squared loss, supplies the PL inequality on $\Rcal$. In our fixed-step theorem, containment is imposed as a hypothesis, and under that hypothesis the PL inequality yields linear convergence on $\Rcal$. The decomposition matters: the LQCR does the localization work, the NTK condition does the PL work, and neither does the other's job.

\section{Main Theoretical Results}
\label{sec:results}

We now state our main theoretical results. The structure follows the decomposition fixed in Section~\ref{sec:framework}: the LQCR supplies a local region with a known radius, while a local condition on the empirical NTK supplies a PL inequality on that region. For fixed-step GD, we impose containment in the region as a hypothesis; combined with the PL inequality, this yields linear convergence while the iterates remain in the region. We state the results for the squared loss, since the bridge from NTK conditioning to the PL inequality used here is a squared-loss identity. We discuss the scope and possible extensions to other losses at the end of the section.

\subsection{Local NTK conditioning}

We replace the uniform spectral assumption used in earlier drafts by a strictly weaker condition: pointwise positivity of the NTK at initialization, together with Lipschitz continuity of the NTK on the region. The uniform positivity needed for the PL inequality is then \emph{derived} via Weyl's inequality, so that the region radius $r(\Rcal)$ and the Lipschitz constant $L_\Theta$ both appear explicitly in the resulting PL constant.

\begin{assumption}[Local NTK conditioning]
\label{assump:ntk}
Let $\Rcal$ be an $(\alpha,\gamma)$-LQCR around an initialization $\theta_0$ in the sense of \citep{Aich2025}, with radius $r(\Rcal)$. We assume that the empirical NTK $\Thetab_\theta = \tfrac{1}{n} J_\theta J_\theta^\top$ satisfies:
\begin{itemize}
    \item \textbf{Pointwise positivity at initialization:} $\lambda_{\min}(\Thetab_{\theta_0}) \ge \lambda_0 > 0.$
    \item \textbf{Lipschitz stability on $\Rcal$:} for all $\theta, \phi \in \Rcal$,
    \begin{align}
        \|\Thetab_\theta - \Thetab_\phi\| \le L_\Theta \|\theta - \phi\|.
    \end{align}
    \item \textbf{Compatibility of region and stability:} $L_\Theta \, r(\Rcal) < \lambda_0.$
\end{itemize}
\end{assumption}

\begin{remark}[On what this assumption asks for]
\label{rem:ntk_assumption}
Pointwise positivity at initialization is a much milder requirement than uniform positivity on all of $\Rcal$, and matches the regime in which NTK-based analyses have been carried out in the literature \citep{Du2019, Liu2020, Oymak2019}. Local Lipschitz stability of the empirical NTK is also a standard regularity property in the finite-width NTK literature \citep{Lee2019, Liu2020}. The compatibility condition $L_\Theta \, r(\Rcal) < \lambda_0$ ties the size of the LQCR to the NTK's regularity: the region must be small enough that NTK drift across it cannot exhaust the initial spectral gap. We do not claim this condition holds in arbitrary practical training; we identify it as the precise local property under which the PL mechanism activates.
\end{remark}

The following lemma is the actual workhorse. It propagates pointwise positivity across the region using Lipschitz stability.

\begin{lemma}[Uniform NTK positivity on $\Rcal$]
\label{lem:weyl}
Under Assumption~\ref{assump:ntk}, for every $\theta \in \Rcal$,
\begin{align}
\lambda_{\min}(\Thetab_\theta) \;\ge\; \lambda_0 - L_\Theta \|\theta - \theta_0\| \;\ge\; \lambda_0 - L_\Theta \, r(\Rcal) \;=:\; \lambda_\Rcal \;>\; 0.
\end{align}
\end{lemma}

\begin{proof}
By Weyl's inequality applied to the symmetric matrices $\Thetab_\theta$ and $\Thetab_{\theta_0}$,
\begin{align}
|\lambda_{\min}(\Thetab_\theta) - \lambda_{\min}(\Thetab_{\theta_0})| \le \|\Thetab_\theta - \Thetab_{\theta_0}\|.
\end{align}
Combined with Lipschitz stability,
\begin{align}
\lambda_{\min}(\Thetab_\theta) \ge \lambda_{\min}(\Thetab_{\theta_0}) - L_\Theta \|\theta - \theta_0\| \ge \lambda_0 - L_\Theta \|\theta - \theta_0\|.
\end{align}
Since $\theta \in \Rcal$, $\|\theta - \theta_0\| \le r(\Rcal)$, so $\lambda_{\min}(\Thetab_\theta) \ge \lambda_0 - L_\Theta r(\Rcal) = \lambda_\Rcal$, which is strictly positive by the compatibility condition.
\end{proof}

Lemma~\ref{lem:weyl} is where the localized region supplied by the LQCR framework enters the analysis. The radius $r(\Rcal)$, supplied by the LQCR construction of \citep{Aich2025} and depending on depth, width, and activation bounds, enters the lower bound on $\lambda_{\min}(\Thetab_\theta)$ directly. Without an explicit region, there is nothing to propagate across; without Lipschitz stability, pointwise positivity cannot be transported off the initialization point. Both ingredients are needed.

\subsection{Existence of LPLRs}

We now show that the LQCR, together with Assumption~\ref{assump:ntk}, is an LPLR for the squared loss, with an explicit PL constant.

\begin{theorem}[Existence of LPLRs under local NTK conditioning, squared loss]
\label{thm:lplr_existence}
Consider the squared empirical loss $\Lcal(\theta) = \frac{1}{2n} \sum_{i=1}^n (f_\theta(x_i) - y_i)^2$. Let $\Rcal$ be an $(\alpha,\gamma)$-LQCR around $\theta_0$ in the sense of \citep{Aich2025}, and suppose Assumption~\ref{assump:ntk} holds on $\Rcal$. Then $\Rcal$ is a $(\mu, \Rcal)$-LPLR with PL constant
\begin{align}
\mu \;=\; \lambda_\Rcal \;=\; \lambda_0 - L_\Theta \, r(\Rcal) \;>\; 0.
\end{align}
\end{theorem}

\begin{proof}
The argument has two steps: a squared-loss identity that relates the gradient norm to the residual via the NTK, and the spectral lower bound of Lemma~\ref{lem:weyl}.

For the squared loss, $\nabla \Lcal(\theta) = \tfrac{1}{n} J_\theta^\top (f_\theta - y)$, so
\begin{align}
\|\nabla \Lcal(\theta)\|^2 = \tfrac{1}{n^2} \|J_\theta^\top (f_\theta - y)\|^2 = \tfrac{1}{n^2} (f_\theta - y)^\top J_\theta J_\theta^\top (f_\theta - y).
\end{align}
By definition of the empirical NTK, $J_\theta J_\theta^\top = n \Thetab_\theta$, so
\begin{align}
\|\nabla \Lcal(\theta)\|^2 = \tfrac{1}{n} (f_\theta - y)^\top \Thetab_\theta (f_\theta - y) \ge \tfrac{\lambda_{\min}(\Thetab_\theta)}{n} \|f_\theta - y\|^2.
\end{align}
By Lemma~\ref{lem:weyl}, for any $\theta \in \Rcal$, $\lambda_{\min}(\Thetab_\theta) \ge \lambda_\Rcal$, so
\begin{align}
\|\nabla \Lcal(\theta)\|^2 \ge \tfrac{\lambda_\Rcal}{n} \|f_\theta - y\|^2.
\label{eq:grad_lb}
\end{align}

For the suboptimality side, observe that for the squared loss,
\begin{align}
\Lcal(\theta) - \Lcal_\Rcal^* \le \Lcal(\theta) = \tfrac{1}{2n} \|f_\theta - y\|^2,
\end{align}
since $\Lcal_\Rcal^* \ge 0$. Rearranged,
\begin{align}
\|f_\theta - y\|^2 \ge 2n \, (\Lcal(\theta) - \Lcal_\Rcal^*).
\label{eq:res_lb}
\end{align}
Substituting \eqref{eq:res_lb} into \eqref{eq:grad_lb} yields
\begin{align}
\|\nabla \Lcal(\theta)\|^2 \ge 2 \lambda_\Rcal \, (\Lcal(\theta) - \Lcal_\Rcal^*),
\end{align}
which is the local PL inequality with constant $\mu = \lambda_\Rcal$.
\end{proof}

\begin{remark}[Where each ingredient enters]
\label{rem:ingredients}
The LQCR enters through the radius $r(\Rcal)$, which controls how far NTK drift can erode the initial spectral gap (Lemma~\ref{lem:weyl}). The Lipschitz constant $L_\Theta$ enters as the rate of that erosion. The initial spectral value $\lambda_0$ enters as the starting budget. The compatibility condition $L_\Theta \, r(\Rcal) < \lambda_0$ guarantees $\mu > 0$. The squared-loss structure is used in two places: $\nabla \Lcal = \tfrac{1}{n} J_\theta^\top (f_\theta - y)$, and the residual bound $\|f_\theta - y\|^2 \ge 2n (\Lcal(\theta) - \Lcal_\Rcal^*)$. The second of these uses $\Lcal_\Rcal^* \ge 0$, which holds for any non-negative loss, but the gradient identity in the first is specific to MSE.
\end{remark}

\begin{remark}[On the tightness of $\mu$]
\label{rem:tightness}
The substitution in \eqref{eq:res_lb} uses $\Lcal_\Rcal^* \ge 0$ rather than $\Lcal_\Rcal^* = 0$. The resulting PL constant $\mu = \lambda_\Rcal$ is therefore tight only in the near-interpolation regime where $\Lcal_\Rcal^* \approx 0$; when $\Lcal_\Rcal^*$ is appreciably positive, $\mu$ is a conservative lower bound. In the empirical setting of Section~\ref{sec:experiments}, increasing width simultaneously increases $\lambda_0$ and decreases $\Lcal_\Rcal^*$ over the moderate-width regime (Table~\ref{tab:width_ablation}), so both factors push the bound toward tightness in that regime.
\end{remark}

\subsection{Linear convergence}

Once the PL inequality is established on $\Rcal$, linear convergence of fixed-step GD on $\Rcal$ is a standard consequence \citep{Polyak1963, Karimi2016}, provided the iterates remain inside $\Rcal$. In the fixed-step statement below, this containment is imposed as a hypothesis rather than derived from \citep{Aich2025}, whose GD containment result uses a decaying step size.

\begin{theorem}[Linear convergence of GD on the LPLR]
\label{thm:linear_conv}
Assume the conditions of Theorem~\ref{thm:lplr_existence}, and additionally that $\Lcal$ is $L$-smooth on $\Rcal$. Consider GD with $\theta^{(t+1)} = \theta^{(t)} - \eta \nabla \Lcal(\theta^{(t)})$ starting from $\theta^{(0)} = \theta_0$. Choose the step size
\begin{align}
\eta \;\le\; \tfrac{1}{L}.
\end{align}
Assume that the resulting iterates remain in $\Rcal$, i.e.,
$\theta^{(t)} \in \Rcal$ for every $t \ge 0$. Then the loss converges linearly to $\Lcal_\Rcal^*$:
\begin{align}
\Lcal(\theta^{(t)}) - \Lcal_\Rcal^* \;\le\; (1 - \eta \lambda_\Rcal)^t \, (\Lcal(\theta_0) - \Lcal_\Rcal^*).
\end{align}
\end{theorem}

\begin{proof}

\textbf{Containment.} The fixed-step linear-rate statement is conditional on the iterates remaining in $\Rcal$. This containment is imposed as part of the theorem hypothesis. Under this containment condition, Assumption~\ref{assump:ntk}, and hence the conclusion of Theorem~\ref{thm:lplr_existence}, applies at every iterate.

\textbf{Linear rate.} By $L$-smoothness, for $\eta \le 1/L$ the descent lemma gives
\begin{align}
\Lcal(\theta^{(t+1)}) \le \Lcal(\theta^{(t)}) - \tfrac{\eta}{2} \|\nabla \Lcal(\theta^{(t)})\|^2.
\end{align}
By Theorem~\ref{thm:lplr_existence}, $\|\nabla \Lcal(\theta^{(t)})\|^2 \ge 2 \lambda_\Rcal (\Lcal(\theta^{(t)}) - \Lcal_\Rcal^*)$. Substituting,
\begin{align}
\Lcal(\theta^{(t+1)}) - \Lcal_\Rcal^* \le (1 - \eta \lambda_\Rcal) (\Lcal(\theta^{(t)}) - \Lcal_\Rcal^*).
\end{align}
Iterating this recursion from $t = 0$ gives the stated rate.
\end{proof}

\begin{remark}[On the step-size requirement]
The fixed-step rate theorem separates descent from containment. The condition $\eta \le 1/L$ is the standard smoothness requirement used in the descent lemma. Containment in $\Rcal$ is imposed as a hypothesis rather than derived from \citep{Aich2025}, whose GD result uses a decaying step size. Without containment, the iterates could in principle leave $\Rcal$, in which case neither the local PL inequality nor the rate is guaranteed.
\end{remark}

\subsection{Scope and limitations}
\label{subsec:scope}

Two scope statements are worth making explicit.

\textbf{Loss function.} Theorem~\ref{thm:lplr_existence} is proved for squared loss. The gradient identity $\nabla \Lcal = \tfrac{1}{n} J_\theta^\top (f_\theta - y)$ used in the proof is specific to MSE, and the residual-suboptimality bound uses non-negativity of the loss. The LPLR definition itself (Definition~\ref{def:lplr_restated}) is loss-agnostic; what is squared-loss specific is the bridge from NTK conditioning to the PL inequality. Extending the bridge to standard classification losses (cross-entropy, logistic) requires a different argument, since the residual-based identity is unavailable. We leave this to future work.

\textbf{Sufficient, not necessary.} Theorem~\ref{thm:lplr_existence} and Theorem~\ref{thm:linear_conv} give sufficient conditions for local PL behavior and linear convergence on the LQCR. They do not claim that the LPLR mechanism is necessary for fast convergence in deep networks, nor that the LQCR/NTK-stability picture is the only mechanism by which linear-rate behavior can arise in practice. Other mechanisms, such as feature-learning dynamics and mean-field analyses, operate under different assumptions and may apply where ours do not. The empirical section is designed to test whether the latent variables of our framework ($\lambda_{\min}(\Thetab_{\theta^{(t)}})$, parameter drift, the PL ratio) actually behave as our theory predicts in the runs where fast convergence is observed; it is not designed to rule out alternative explanations, and we do not claim it does.
\section{Empirical Validation}
\label{sec:experiments}

We design experiments that probe the four latent variables of Theorem~\ref{thm:lplr_existence} and Theorem~\ref{thm:linear_conv} directly, rather than only reporting loss curves. Specifically, along the training trajectory we track: (i) the subset empirical NTK eigenvalue $\lambda_{\min}(\Thetab_{\theta^{(t)}})$ on a fixed input subset, which probes Lemma~\ref{lem:weyl}; (ii) the parameter drift $\|\theta^{(t)} - \theta_0\|$, which probes the containment hypothesis used in Theorem~\ref{thm:linear_conv}; (iii) the empirical PL ratio $\|\nabla\Lcal(\theta^{(t)})\|^2 / [2(\Lcal(\theta^{(t)}) - \widehat{\Lcal}_\Rcal^*)]$, which directly tests the local PL inequality with constant $\mu = \lambda_\Rcal$; and (iv) the suboptimality gap $\Lcal(\theta^{(t)}) - \widehat{\Lcal}_\Rcal^*$ on a semi-log scale, which tests the linear-rate conclusion. The regional minimum is proxied by the final training loss, $\widehat{\Lcal}_\Rcal^* := \Lcal(\theta^{(T)})$; we discuss the implications of this proxy where relevant. All experiments use the squared loss, consistent with the scope of Theorem~\ref{thm:lplr_existence}.

\subsection{Controlled validation on binary MNIST}
\label{subsec:exp_mnist}

\textbf{Setup.} We train a fully connected network with five affine layers: four hidden layers of width $512$ followed by a scalar output layer with smooth Softplus activations ($\beta = 1$) on a binary subset of MNIST (digits 3 vs.\ 8, $n = 11{,}982$ training samples). We use full-batch gradient descent with squared loss, He initialization, and step size $\eta = 10^{-3}$, run for $T = 250$ epochs. The smooth activation is chosen to match the assumptions of \citep{Aich2025}; the small width and full-batch setting are chosen so the four diagnostics above can be measured precisely. The subset empirical NTK is computed on a fixed random subset of $n_{\text{sub}} = 100$ training inputs (held constant across all probes), as $\Thetab_{\theta^{(t)}}^{\text{sub}} = \tfrac{1}{n_{\text{sub}}} J_{\theta^{(t)}}^{\text{sub}} (J_{\theta^{(t)}}^{\text{sub}})^\top$, and probed every $5$ epochs.

\begin{figure*}[t]
    \centering
    \begin{subfigure}[t]{0.48\textwidth}
        \centering
        \includegraphics[width=\linewidth]{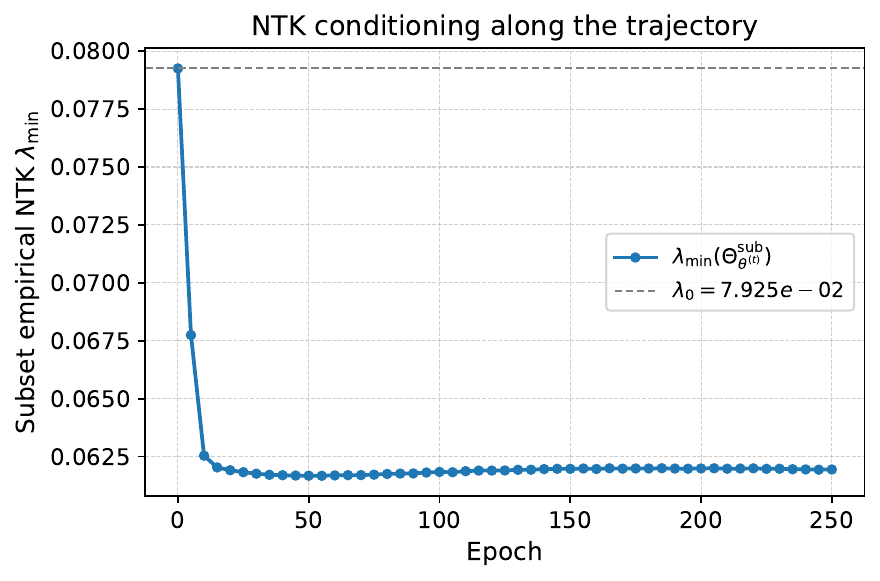}
        \caption{Subset empirical NTK eigenvalue.}
        \label{fig:exp1_ntk}
    \end{subfigure}
    \hfill
    \begin{subfigure}[t]{0.48\textwidth}
        \centering
        \includegraphics[width=\linewidth]{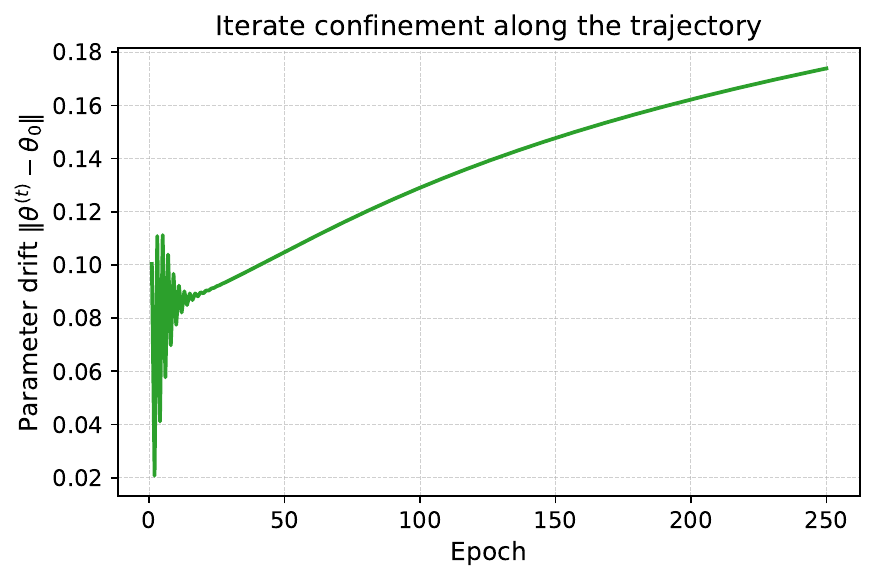}
        \caption{Parameter drift from initialization.}
        \label{fig:exp1_drift}
    \end{subfigure}

    \vspace{0.5em}

    \begin{subfigure}[t]{0.48\textwidth}
        \centering
        \includegraphics[width=\linewidth]{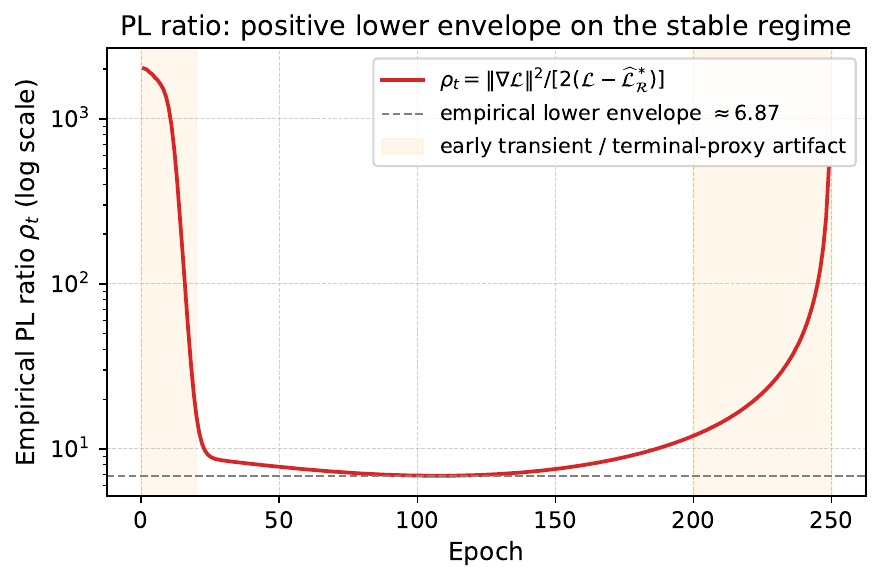}
        \caption{Empirical PL ratio.}
        \label{fig:exp1_pl}
    \end{subfigure}
    \hfill
    \begin{subfigure}[t]{0.48\textwidth}
        \centering
        \includegraphics[width=\linewidth]{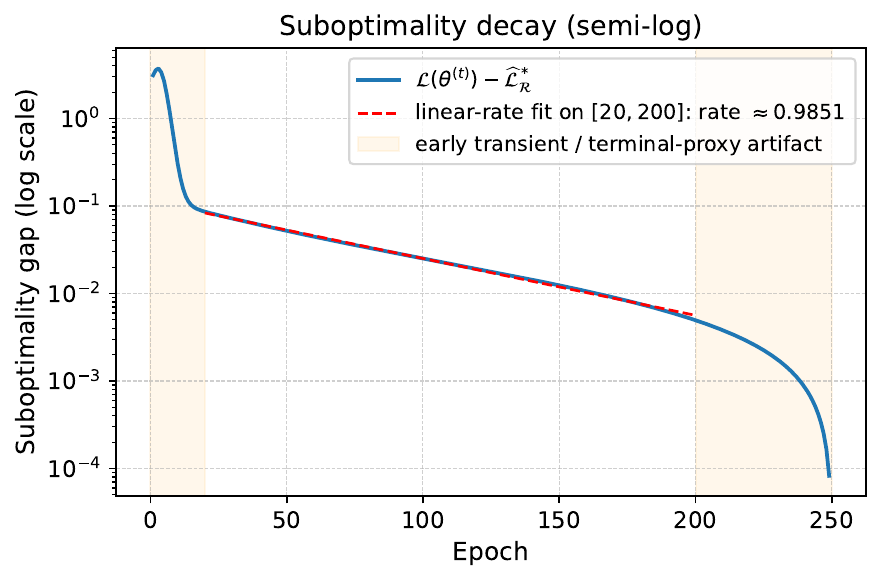}
        \caption{Suboptimality decay on a semi-log scale.}
        \label{fig:exp1_loss}
    \end{subfigure}

    \caption{Controlled binary MNIST validation. The diagnostics track the latent variables in Theorems~\ref{thm:lplr_existence}--\ref{thm:linear_conv}: subset empirical NTK conditioning, parameter drift, empirical PL ratio, and suboptimality decay. The shaded regions in panels (c) and (d) mark the early transient and terminal-loss-proxy affected regimes.}
    \label{fig:exp1_all}
\end{figure*}

Figure~\ref{fig:exp1_all} summarizes the four diagnostics used in this controlled validation experiment.

\textbf{NTK conditioning along the trajectory.} The subset empirical NTK $\lambda_{\min}$ drops from $\lambda_0 = 7.93 \times 10^{-2}$ at initialization to approximately $6.2 \times 10^{-2}$ over the first $\sim 15$ epochs, after which it stabilizes on a plateau and remains essentially flat for the remaining $235$ epochs, ending at $\lambda_T = 6.19 \times 10^{-2}$ (Figure~\ref{fig:exp1_ntk}). The total drift is approximately $22\%$ of the initial spectral gap, concentrated almost entirely in the early phase. The kernel remains positive throughout. This pattern is consistent with the conclusion of Lemma~\ref{lem:weyl}: pointwise positivity at initialization is preserved along the training trajectory, leaving a positive $\lambda_\Rcal$ that the GD dynamics can exploit for the bulk of the run.

\textbf{Iterate confinement.} The parameter drift $\|\theta^{(t)} - \theta_0\|$ exhibits an initial oscillatory transient over the first $\sim 15$ epochs as the optimizer adjusts from initialization, then settles into smooth monotone growth, reaching a final value of $0.174$ (Figure~\ref{fig:exp1_drift}). For a network with on the order of $10^6$ parameters, this drift is small relative to the dimension of the parameter space, and is consistent with the regime in which Assumption~\ref{assump:ntk}'s Lipschitz stability is most meaningful. The iterates remain in a small neighborhood of initialization, consistent with the containment hypothesis in Theorem~\ref{thm:linear_conv}.

\textbf{Empirical PL ratio and a transparency note on the proxy.} We track the empirical PL ratio
\begin{align}
\rho_t \;:=\; \frac{\|\nabla\Lcal(\theta^{(t)})\|^2}{2(\Lcal(\theta^{(t)}) - \widehat{\Lcal}_\Rcal^*)}
\end{align}
along training (Figure~\ref{fig:exp1_pl}). Two regimes near the endpoints of training are artifacts of the terminal-loss proxy $\widehat{\Lcal}_\Rcal^* := \Lcal(\theta^{(T)})$ rather than properties of the underlying dynamics: very early epochs ($t \lesssim 15$) where the gradient norm is dominated by the warm-up transient, and very late epochs ($t \gtrsim 230$) where the denominator $(\Lcal(\theta^{(t)}) - \widehat{\Lcal}_\Rcal^*)$ tends to zero by construction. We therefore focus the assessment on the stable middle regime $20 \le t \le 200$, on which $\rho_t$ is bounded below by approximately $6.87$ and varies slowly. The existence of a positive lower envelope on this regime is a direct test of the local PL inequality predicted by Theorem~\ref{thm:lplr_existence}: in contrast to a fitted log-log slope, it is the uniform one-sided lower bound $\rho_t \ge \mu$ that the theorem actually asserts.

\textbf{Comparison to the theoretical bound.} The empirical PL ratio on the stable regime is substantially larger than the conservative theoretical lower bound $\mu = \lambda_\Rcal \approx 6 \times 10^{-2}$. This is consistent with Remark~\ref{rem:tightness}: the bound $\mu = \lambda_\Rcal$ is derived using $\widehat{\Lcal}_\Rcal^* \ge 0$ rather than $\widehat{\Lcal}_\Rcal^* = 0$, and is tight only in the near-interpolation regime. The final training loss here is $\widehat{\Lcal}_\Rcal^* \approx 0.049$, which is not near zero, so a sizable gap between the empirical PL ratio and the theoretical lower bound is expected. The width ablation in Section~\ref{subsec:exp_width} examines how this gap changes with width.

\textbf{Suboptimality decay.} The suboptimality gap $\Lcal(\theta^{(t)}) - \widehat{\Lcal}_\Rcal^*$ exhibits three distinct phases on the semi-log plot (Figure~\ref{fig:exp1_loss}): an initial transient over the first $\sim 15$ epochs, a long regime of approximately linear decay from epoch $\sim 20$ to epoch $\sim 200$, and a sharp acceleration over the final $\sim 30$ epochs that reflects the terminal-loss proxy and not a genuine super-linear regime. A linear-rate fit on the middle regime $[20, 200]$ yields a contraction factor of approximately $0.9851$, supporting the qualitative geometric-decay behavior predicted by Theorem~\ref{thm:linear_conv}. The qualitative shape of the loss curve over the bulk of training matches the linear-rate prediction of the theorem.

\textbf{Summary.} On the stable middle regime $[20, 200]$, the four diagnostics behave as Theorems~\ref{thm:lplr_existence}--\ref{thm:linear_conv} predict: the NTK is positive and approximately constant, the iterates grow slowly within a small neighborhood of initialization, the empirical PL ratio admits a positive lower envelope ($\rho_t \ge 6.87$), and the suboptimality decays geometrically. The remaining behavior near the endpoints is honestly attributable to the early warm-up transient and to the terminal-loss proxy for $\widehat{\Lcal}_\Rcal^*$. Together these diagnostics test the latent variables of the LPLR mechanism in a single controlled run, rather than its consequences alone.

\subsection{Width ablation and the role of the step-size condition}
\label{subsec:exp_width}

\textbf{Setup.} We repeat the controlled MNIST experiment of Section~\ref{subsec:exp_mnist} for four widths $m \in \{128, 256, 512, 1024\}$, holding all other settings fixed (depth $5$, Softplus activations, full-batch GD, $\eta = 10^{-3}$, $T = 250$ epochs, He init, same seed across runs). The NTK is probed on the same fixed $n_{\text{sub}} = 100$ subset across all four runs, so $\lambda_{\min}(\Thetab^{\text{sub}})$ values are directly comparable. The intent is to test the width-dependent predictions of Theorem~\ref{thm:lplr_existence} and Lemma~\ref{lem:weyl}, and to map the regime in which those predictions hold.

\textbf{Observed moderate-width pattern at widths $128$--$512$.} Table~\ref{tab:width_ablation} summarizes the key quantities. The first three widths show a stable local-regime pattern:
\begin{enumerate}
    \item $\lambda_0$ grows monotonically with width in these runs, approximately doubling between successive widths ($2.2 \times 10^{-2}$, $4.2 \times 10^{-2}$, $7.9 \times 10^{-2}$).
    \item NTK drift $|\lambda_0 - \lambda_T|$ remains small relative to $\lambda_0$ in each case (drift $\le 22\%$ of $\lambda_0$), so the observed NTK behavior is consistent with the LQCR compatibility condition $L_\Theta \, r(\Rcal) < \lambda_0$ from Assumption~\ref{assump:ntk}, and the subset empirical NTK remains positive throughout training.
    \item The terminal-loss proxy $\widehat{\Lcal}_\Rcal^*$ decreases monotonically with width ($0.066$, $0.058$, $0.049$), consistent with wider networks reaching lower-loss regions.
    \item Parameter drift $\|\theta^{(T)} - \theta_0\|$ stays in a narrow range around $0.15$--$0.18$ across the three widths, indicating that iterates remain in a small neighborhood of initialization (the lazy regime).
    \item The empirical PL ratio on the stable middle regime stays in the same order of magnitude across these widths ($5.1$, $7.0$, $6.9$).
\end{enumerate}

\begin{table}[ht]
\centering
\caption{Width ablation on binary MNIST (full-batch GD, 250 epochs). NTK probed on a fixed $n_{\text{sub}} = 100$ subset; PL ratio computed on the stable middle regime $[20, 200]$ identified in Section~\ref{subsec:exp_mnist}. Top block: widths $128$--$512$ at $\eta = 10^{-3}$ show stable moderate-width local-regime behavior. Middle block: width $1024$ at $\eta = 10^{-3}$ hits the boundary of the local regime. Bottom block: width $1024$ at $\eta = 5 \times 10^{-4}$ recovers the lazy regime and yields the largest empirical PL ratio of the study.}
\label{tab:width_ablation}
\small
\begin{tabular}{r r r r r r r r}
\toprule
Width ($\eta$) & \# params & $\lambda_0$ & $\lambda_T$ & NTK drift & $\widehat{\Lcal}_\Rcal^*$ & PL ratio (min) & $\|\theta^{(T)}\!-\!\theta_0\|$ \\
\midrule
$128$ ($10^{-3}$) & 150{,}145 & $2.20\!\times\!10^{-2}$ & $2.12\!\times\!10^{-2}$ & $7.59\!\times\!10^{-4}$ & $6.61\!\times\!10^{-2}$ & $5.12$ & $0.153$ \\
$256$ ($10^{-3}$) & 398{,}593 & $4.16\!\times\!10^{-2}$ & $4.10\!\times\!10^{-2}$ & $6.12\!\times\!10^{-4}$ & $5.79\!\times\!10^{-2}$ & $7.05$ & $0.178$ \\
$512$ ($10^{-3}$) & 1{,}190{,}401 & $7.93\!\times\!10^{-2}$ & $6.19\!\times\!10^{-2}$ & $1.73\!\times\!10^{-2}$ & $4.89\!\times\!10^{-2}$ & $6.87$ & $0.174$ \\
\midrule
$1024$ ($10^{-3}$) & 3{,}953{,}665 & $1.48\!\times\!10^{-1}$ & $2.49\!\times\!10^{-2}$ & $1.23\!\times\!10^{-1}$ & $6.52\!\times\!10^{-2}$ & $4.60$ & $1.870$ \\
\midrule
$1024$ ($5\!\times\!10^{-4}$) & 3{,}953{,}665 & $1.48\!\times\!10^{-1}$ & $1.09\!\times\!10^{-1}$ & $3.90\!\times\!10^{-2}$ & $5.30\!\times\!10^{-2}$ & $14.39$ & $0.158$ \\
\bottomrule
\end{tabular}
\end{table}

\textbf{The boundary at $(m, \eta) = (1024, 10^{-3})$.} The widest configuration at the same step size used at narrower widths behaves qualitatively differently. The parameter drift $\|\theta^{(T)} - \theta_0\|$ jumps to $1.87$ within the first $\sim 50$ epochs and then stays essentially constant; the NTK $\lambda_{\min}$ decays from $\lambda_0 = 1.48 \times 10^{-1}$ to $\lambda_T = 2.49 \times 10^{-2}$, a drop of approximately $83\%$ of the initial spectral gap; the empirical PL ratio drops to $4.60$. The model still trains and reaches a comparable final loss, but the LQCR compatibility condition $L_\Theta \, r(\Rcal) < \lambda_0$ is no longer plausibly maintained across this trajectory (we do not compute $L_\Theta$ or $r(\Rcal)$ numerically; we read this off the observed NTK collapse and drift); the iterates have left the lazy regime in which our framework's predictions are tightest.

This is consistent with the fixed-step scope of Theorem~\ref{thm:linear_conv}: the descent argument requires $\eta \le 1/L$, while containment in $\Rcal$ is a separate hypothesis. A step size that appears to preserve the local-regime diagnostics at $m = 512$ need not preserve them at $m = 1024$.

\textbf{Recovery at $(m, \eta) = (1024, 5 \times 10^{-4})$.} To test whether the boundary behavior above is primarily step-size driven rather than an unavoidable consequence of width, we re-run the width-$1024$ configuration with a reduced step size $\eta = 5 \times 10^{-4}$. All other settings (depth, activation, initialization, dataset, epochs, NTK probe protocol, random seed) are held fixed. The reduced step size restores the lazy regime cleanly across all four diagnostics, as the bottom block of Table~\ref{tab:width_ablation} shows.

Four observations summarize the recovery. First, the parameter drift drops from $1.87$ to $0.158$, a reduction of roughly $12\times$, returning the iterates to a comparably small parameter-drift regime observed at widths $128$--$512$. Second, the NTK drift falls from $83\%$ of $\lambda_0$ to $26\%$ of $\lambda_0$, comparable to the drift observed at $m = 512$ with $\eta = 10^{-3}$, so the observed NTK behavior is consistent with remaining in the local regime along the trajectory. Third, the terminal-loss proxy $\widehat{\Lcal}_\Rcal^*$ also decreases ($0.065 \to 0.053$), so the reduced step size yields a strictly better terminal loss in addition to a more theory-consistent trajectory. Fourth, and most directly tied to the theory, the empirical PL ratio on the stable middle regime increases to $14.4$, the largest value in the entire ablation, consistent with the largest $\lambda_0$ in the study being preserved rather than eroded.

\textbf{Takeaway.} The ablation supports the predicted NTK-conditioning trend at moderate widths and identifies the boundary beyond which a fixed step size fails to maintain the local regime; the recovery indicates that the boundary behavior is at least partly step-size driven rather than an unavoidable consequence of width, and the largest-width run produces the largest empirical PL constant once the reduced step size restores the observed local-regime diagnostics. The sharper, scope-aware reading is what our theorems actually claim, and is more informative than a simple monotone ``wider is better" statement on final loss.

\subsection{Robustness check on a CNN under SGD with cosine annealing}
\label{subsec:exp_cnn}

\textbf{Scope of this experiment.} The previous subsections test the LPLR mechanism in the setting where the theory is most directly applicable: a fully-connected network with smooth activations under full-batch gradient descent and a fixed step size. This subsection asks a different question: do the diagnostic signatures of the LPLR mechanism --- a positive lower envelope on the empirical PL ratio, bounded parameter drift, and approximately linear-rate suboptimality decay --- persist when we move outside that controlled setting into a deeper convolutional architecture trained with mini-batch SGD and a cosine learning-rate schedule? This is a robustness check, not a direct test of the mechanism: NTK eigenvalue tracking is not feasible at this parameter scale, so $\lambda_{\min}(\Thetab_{\theta^{(t)}})$ is not directly verified. We measure only the three diagnostics that remain tractable.

\textbf{Setup.} We train a ResNet-style CNN \citep{he2016deep} with three stages of two residual blocks each, base width $32$ (so channel widths are $32/64/128$), and \emph{GroupNorm} \citep{wu2018group} (groups $=8$) on a 5-class subset of CIFAR-10 \citep{krizhevsky2009learning} ($n = 25{,}000$ training samples). GroupNorm is used rather than BatchNorm \citep{ioffe2015batch} so that the full-dataset diagnostic objective is not coupled to mini-batch normalization statistics. The model has $736{,}293$ parameters. Training uses squared loss on one-hot targets, SGD with momentum $0.9$, cosine-annealing schedule with base learning rate $10^{-3}$, batch size $128$, no weight decay, for $T = 200$ epochs. We repeat the run for three random seeds $\{42, 43, 44\}$. Because this setting uses shuffled mini-batch SGD and nondeterministic GPU convolution kernels, the per-seed PL-ratio minimum varies modestly between reruns even at fixed seeds; what is stable across reruns, and what the theorem actually concerns, is the existence of a positive lower envelope on the stable regime rather than any single numerical value. Diagnostics (full-dataset $\Lcal(\theta^{(t)})$ and full-dataset $\|\nabla\Lcal(\theta^{(t)})\|^2$) are computed every epoch via micro-batch gradient accumulation on the entire training set, so the diagnostics are exact rather than noisy mini-batch estimates. PL ratio statistics are computed on a stable middle regime $[20, 180]$ analogous to the one identified in Section~\ref{subsec:exp_mnist}.

\begin{figure*}[t]
    \centering
    \begin{subfigure}[t]{0.32\textwidth}
        \centering
        \includegraphics[width=\linewidth]{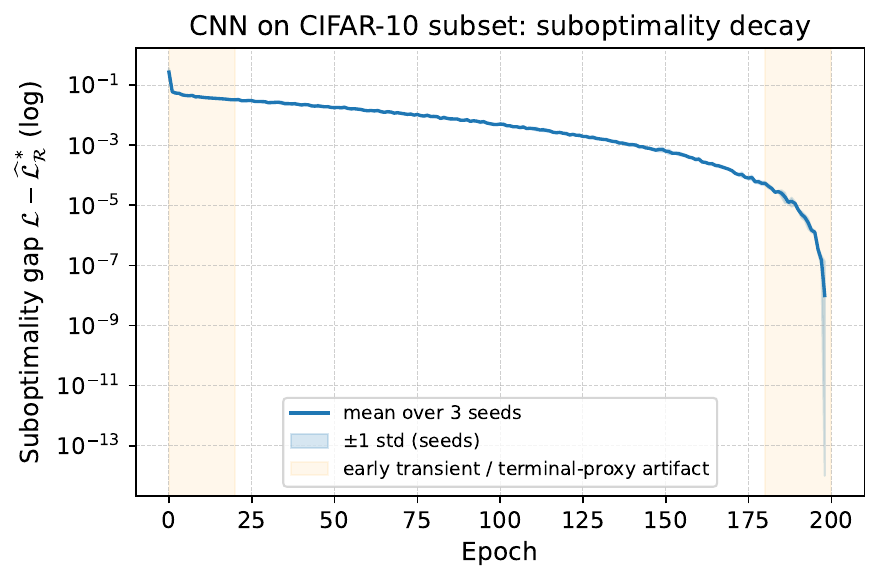}
        \caption{Suboptimality decay.}
        \label{fig:exp3_loss}
    \end{subfigure}
    \hfill
    \begin{subfigure}[t]{0.32\textwidth}
        \centering
        \includegraphics[width=\linewidth]{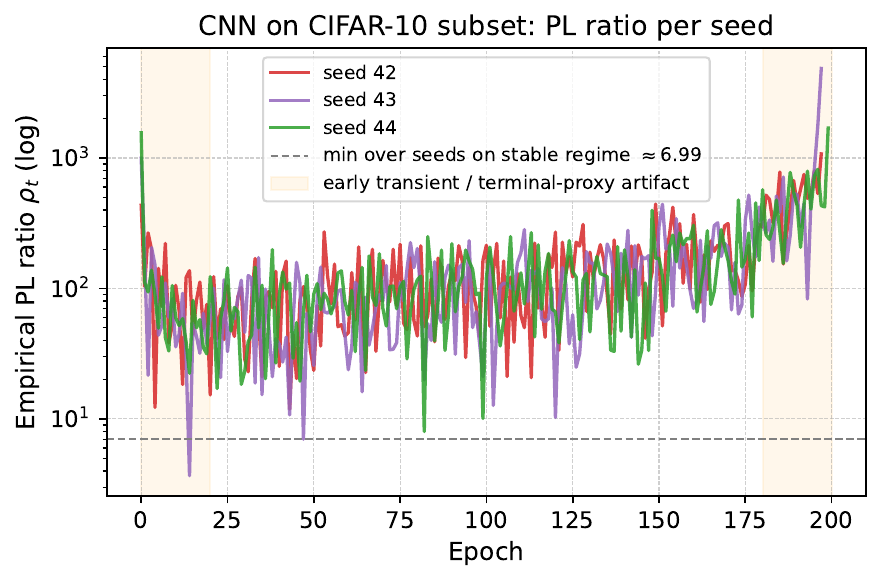}
        \caption{Empirical PL ratio.}
        \label{fig:exp3_pl}
    \end{subfigure}
    \hfill
    \begin{subfigure}[t]{0.32\textwidth}
        \centering
        \includegraphics[width=\linewidth]{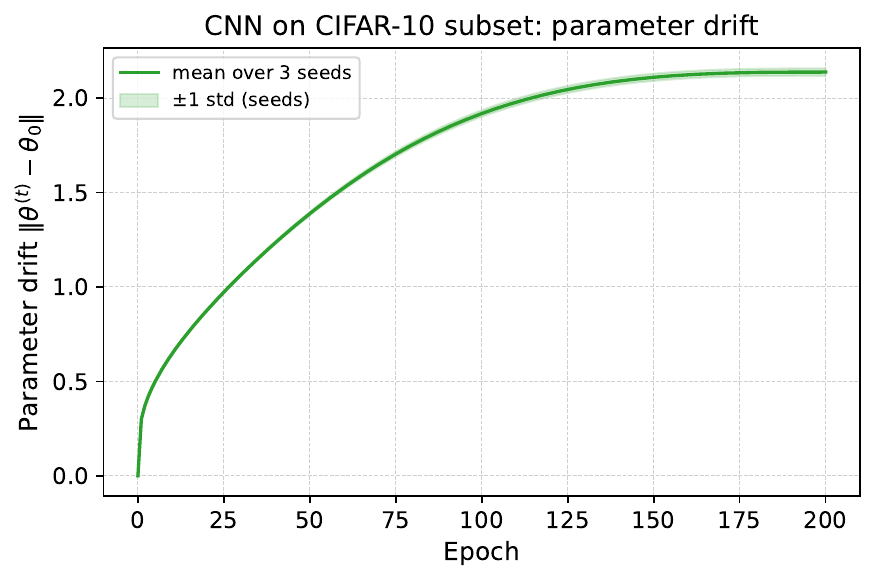}
        \caption{Parameter drift.}
        \label{fig:exp3_drift}
    \end{subfigure}

    \caption{CNN robustness check on a 5-class CIFAR-10 subset. The plots show mean suboptimality decay, per-seed empirical PL ratios, and parameter drift across three seeds. Shaded regions mark early transient and terminal-loss-proxy affected regimes.}
    \label{fig:exp3_all}
\end{figure*}

\textbf{Cross-seed summary.} Table~\ref{tab:exp3_cnn_summary} reports per-seed and mean values of the three diagnostics.

\begin{table}[ht]
\centering
\caption{CNN on CIFAR-10 subset, three seeds. PL ratio on the stable regime $[20, 180]$ is computed per seed using each seed's own terminal-loss proxy $\widehat{\Lcal}_\Rcal^{*}$. Across all three seeds the empirical PL ratio admits a positive lower envelope on the stable regime, and the final parameter drift is tightly clustered.}
\label{tab:exp3_cnn_summary}
\small
\begin{tabular}{r r r r r}
\toprule
Seed & $\widehat{\Lcal}_\Rcal^{*}$ & $\|\theta^{(T)}\!-\!\theta_0\|$ & PL min on $[20,180]$ & PL median on $[20,180]$ \\
\midrule
$42$  & $8.54\!\times\!10^{-3}$ & $2.11$ & $11.89$ & $111.2$ \\
$43$  & $8.30\!\times\!10^{-3}$ & $2.15$ & $6.99$  & $93.6$  \\
$44$  & $8.38\!\times\!10^{-3}$ & $2.15$ & $8.01$  & $101.9$ \\
\midrule
mean  & $8.41\!\times\!10^{-3}$ & $2.14$ & $8.96$  & $102.2$ \\
\bottomrule
\end{tabular}
\end{table}

\textbf{Suboptimality decay.} Figure~\ref{fig:exp3_loss} shows the per-seed mean suboptimality gap with a $\pm 1$ standard deviation band across seeds. On the stable middle regime $[20, 180]$, the suboptimality decays nearly linearly on the semi-log scale over approximately four orders of magnitude, with a very narrow cross-seed band indicating tight reproducibility of the dynamics. The acceleration in the last $\sim 20$ epochs is the terminal-loss proxy artifact in pure form (by construction $\Lcal^{(s)}(\theta^{(T)}) - \widehat{\Lcal}_\Rcal^{*,(s)} = 0$ for each seed) and should not be interpreted as a genuine super-linear regime; the shaded region indicates this range. The overall shape on the stable regime is qualitatively consistent with the linear-rate prediction of Theorem~\ref{thm:linear_conv}.

\textbf{PL ratio.} Figure~\ref{fig:exp3_pl} plots the per-seed PL ratio on a log scale. On the stable middle regime $[20, 180]$, all three seeds maintain a positive lower envelope: the minimum PL ratio across all three seeds on the stable regime is approximately $7.0$, and the median across the three seeds and the stable epochs is approximately $102$. The trajectories show high-frequency oscillation typical of SGD with a time-varying learning rate, including occasional pointwise dips on individual seeds. What the theorem predicts is the existence of a positive lower bound $\rho_t \ge \mu$, i.e.\ a positive envelope, not pointwise tightness; the envelope is comfortably positive throughout the stable regime for every seed.

\textbf{On the tightness of the bound.} The per-seed terminal-loss proxy is $\widehat{\Lcal}_\Rcal^{*} \approx 8.4 \times 10^{-3}$, two orders of magnitude smaller than in the binary-MNIST experiments of Section~\ref{subsec:exp_mnist}. This puts the CNN closer to the near-interpolation regime in which the bound $\mu = \lambda_\Rcal$ of Remark~\ref{rem:tightness} is tightest. The large median PL ratio ($\approx 102$) should be interpreted cautiously because the terminal-loss proxy can inflate the ratio near late training. Nevertheless, on the stable middle regime $[20,180]$, the ratio remains uniformly positive across all seeds, which is the qualitative PL-type signature tested here.

\textbf{Parameter drift.} Figure~\ref{fig:exp3_drift} shows the parameter drift across seeds. The drift grows monotonically but saturates around epoch $\sim 175$ at a value of $\approx 2.14$, with a near-invisible cross-seed standard deviation band, indicating that the optimizer settles into a well-defined local neighborhood that is consistent across initializations. In absolute terms this drift is larger than the drift observed in the controlled MLP setting ($\sim 0.17$ at width $512$). We do not claim the CNN setting falls inside the same lazy regime as the MLP. What we do observe is that the drift is bounded, saturating, monotone over the stable regime, and reproducible across seeds, which is consistent with the iterates remaining inside \emph{some} local region throughout training, even if we cannot certify that region against the LQCR construction directly at this scale.

\textbf{Takeaway.} The three tractable diagnostic signatures associated with the LPLR picture, a positive PL ratio envelope, an approximately linear-rate decay on the bulk of training, and bounded reproducible parameter drift, persist across three independent runs of a ResNet-style CNN trained on a CIFAR-10 subset with mini-batch SGD and cosine annealing. We emphasize that this is a robustness check rather than a direct mechanism test: NTK conditioning is not verified at this parameter scale, the cosine schedule moves outside the fixed-step-size scope of Theorem~\ref{thm:linear_conv}, and the CNN architecture is not the smooth fully-connected setting in which the LQCR construction of \citep{Aich2025} is established. The signatures nevertheless survive the move to a more realistic training pipeline, which we read as evidence that the LPLR mechanism is not confined to the controlled settings of Sections~\ref{subsec:exp_mnist}--\ref{subsec:exp_width}, while remaining careful not to claim that the LPLR mechanism is the unique or even dominant explanation of the observed behavior in this more realistic setting.

\section{Discussion and Conclusion}
\label{sec:conclusion}

This article revisits a question motivated by the prior LQCR framework of \citep{Aich2025}: within a local region of parameter space where gradient descent admits containment and a sublinear stationarity guarantee under the prescribed decaying step-size schedule, can a stronger local rate be obtained under additional local structure? Our answer is conditional. Under a local Neural Tangent Kernel conditioning assumption, namely pointwise positivity at initialization together with Lipschitz continuity over the LQCR, with a compatibility condition tying region radius to the initial spectral gap, we show that the squared empirical loss satisfies a Polyak--{\L}ojasiewicz inequality on the LQCR, with PL constant $\mu = \lambda_\Rcal = \lambda_0 - L_\Theta r(\Rcal)$. For fixed-step GD, our linear-rate theorem assumes that the iterates remain inside the LQCR and then obtains linear convergence on that region. We call such a region a \emph{Locally Polyak--{\L}ojasiewicz Region}.

\textbf{What the framework does and does not claim.} The LPLR construction is a sufficient-condition result, scoped to the squared loss and to the local regime where the iterates remain inside an LQCR satisfying the NTK conditioning assumption. The LQCR plays a specific role: it supplies a localized region with an explicit radius $r(\Rcal)$ and, in the original LQCR result of \citep{Aich2025}, an iterate-confinement guarantee under a decaying step-size schedule. The PL inequality itself is derived from NTK conditioning under squared loss via Weyl's inequality (Lemma~\ref{lem:weyl}), and is not a consequence of the LQCR curvature or descent conditions alone. We do not claim that the LPLR mechanism is necessary for fast convergence in deep networks, nor that it is the only route by which linear-rate behavior can arise in practice. Other mechanisms, such as feature-learning dynamics and mean-field analyses, operate under different assumptions and may apply where ours do not.

\textbf{Relation to prior work.} Linear convergence of GD under NTK, Gram-matrix, or tangent-kernel conditioning is a well-established mechanism going back to \citep{Du2019, AllenZhu2019, Oymak2019}. Those results typically obtain global or trajectory-wide guarantees through overparameterization and kernel or Gram-matrix conditioning. Our contribution is to formulate the same PL mechanism locally, inside a geometrically characterized finite-width region, from a pointwise spectral condition at initialization combined with Lipschitz stability across the region. The result is not a global replacement for those analyses: it trades global guarantees for local hypotheses on a specific region. Architectural approaches such as PLNet \citep{wang2024monotone} build PL-like behavior into specialized networks by construction; our analysis is complementary in addressing standard architectures via a local condition. Stochastic SGD analyses under local {\L}ojasiewicz-type conditions \citep{an2024convergence} and recent local-PL results for two-layer linear networks \citep{xu2025local} share the local-condition philosophy; our framework instantiates it concretely for nonlinear finite-width networks via the LQCR/NTK pair.

\textbf{What the experiments tested and showed.} Section~\ref{sec:experiments} departs from the prior approach of reporting only loss curves and instead probes the latent variables of the theory directly. In a controlled MNIST setting (Section~\ref{subsec:exp_mnist}), the subset empirical NTK $\lambda_{\min}$ remains positive throughout training and stabilizes on a plateau after a short initial drift; the parameter drift remains small relative to the dimension of the parameter space; the empirical PL ratio admits a positive lower envelope on a clearly identifiable middle regime; and the suboptimality gap decays approximately linearly on the semi-log scale on that same regime. The width ablation (Section~\ref{subsec:exp_width}) shows stable moderate-width patterns in $\lambda_0$ and NTK drift at $m \in \{128, 256, 512\}$, exhibits a boundary case at $(m, \eta) = (1024, 10^{-3})$ where the local-regime diagnostics degrade, and recovers the profile at $(m, \eta) = (1024, 5 \times 10^{-4})$ with the largest empirical PL constant of the study, consistent with the fixed-step theorem's separation between descent and containment. A robustness check on a ResNet-style CNN under mini-batch SGD with cosine annealing (Section~\ref{subsec:exp_cnn}) finds that the three tractable diagnostics (PL envelope, linear-rate bulk decay, bounded reproducible drift) persist across three independent seeds, even though NTK conditioning cannot be directly verified at that parameter scale and the schedule moves outside the fixed-step-size scope of Theorem~\ref{thm:linear_conv}.

\textbf{Proper reading of the empirical evidence.} Two interpretive points deserve emphasis. First, our diagnostics test the latent variables of the framework directly, not only its consequences.   This is a stronger empirical test than fitting a log-log slope to a loss curve, but it is still a sufficient-condition test --- it does not call into question alternative explanations for fast convergence. Second, the terminal-loss proxy $\widehat{\Lcal}_\Rcal^*$ used in all experiments compresses the dynamics near the end of training by construction; we identify a stable middle regime in each experiment and restrict the quantitative readouts to that regime, with the proxy-affected endpoints flagged explicitly in plots and text.

\textbf{Limitations and scope.}
\begin{enumerate}
    \item \emph{Loss function.} The PL bridge used in Theorem~\ref{thm:lplr_existence} is a squared-loss identity. The LPLR concept itself is loss-agnostic, but the bridge argument does not transfer directly to classification losses such as cross-entropy, which lack the residual-based gradient identity. Extending the PL derivation to other losses is open.
    \item \emph{Optimizer.} Theorem~\ref{thm:linear_conv} is stated for full-batch GD with a fixed step size. A rigorous extension to mini-batch SGD would need to quantify convergence to a noise ball and characterize how stochasticity interacts with remaining within the LPLR; the empirical CNN result suggests this is plausible but is not a proof.
    \item \emph{Architectural scope.} The LQCR construction of \citep{Aich2025} is established for smooth fully-connected networks. Extending the same construction to architectures with non-smooth activations (e.g.\ ReLU) via subgradient analysis, to residual blocks, or to attention layers, would broaden the applicability of the LPLR framework.
    \item \emph{Step-size adaptation.} The width-ablation boundary at $(m, \eta) = (1024, 10^{-3})$ illustrates that step size affects whether the observed local-regime diagnostics are preserved when width changes. Since the fixed-step theorem assumes containment rather than deriving it, practical schedules that maintain containment or stable local-regime diagnostics across widths would be a useful follow-up.
    
    \item \emph{NTK conditioning at scale.} We verify the NTK conditioning assumption directly only in the MLP setting, where subset empirical NTK eigenvalues are tractable. Establishing or refuting analogous conditioning at the scale of practical CNNs and transformers remains an open empirical question.
\end{enumerate}

\textbf{Conclusion.} The Locally Polyak--{\L}ojasiewicz Region provides one concrete local sufficient condition under which gradient descent on the squared loss converges linearly in a finite-width setting. The framework decomposes cleanly: the LQCR of \citep{Aich2025} supplies the local region and, in its original decaying-step result, a containment mechanism; our fixed-step theorem assumes containment on that region; local NTK conditioning supplies the PL inequality; and the combination yields a rate. The empirical diagnostics introduced here probe each component of this decomposition rather than its end-to-end consequence, and find behavior consistent with the theory on the regimes where the theory applies, including a sharp empirical illustration of where the local-regime requirement breaks and recovers. We hope the framework, and especially the diagnostic style of probing latent variables rather than fitting slopes, will be useful in future work that asks when and why fast optimization arises in finite-width neural networks.

\section*{Broader Impact Statement}

This work is theoretical and methodological in nature. Its primary aim is to improve understanding of when local linear convergence can be justified for gradient descent in finite-width neural networks. A potential positive impact is that sharper optimization diagnostics may help researchers better assess when training dynamics are stable and when step-size choices push the iterates outside the local regime where the local hypotheses of the theory are expected to apply. We do not introduce a deployed system, dataset, or application-specific model, and we do not make claims about direct societal deployment. As with other theoretical work in deep learning optimization, any downstream impact depends on how the resulting methods or diagnostics are used in practical systems.

\bibliography{references}

@article{Aich2025,
  author    = {Aich, Agnideep and Aich, Ashit Baran and Wade, Bruce},
  title     = {Convergence Guarantees for Gradient Descent in Deep Neural Networks with Non-convex Loss Functions},
  journal   = {International Journal of Computer Mathematics},
  year      = {2025},
  volume    = {102},
  number    = {11},
  pages     = {1808--1823},
  doi       = {10.1080/00207160.2025.2522349}
}

@inproceedings{Jacot2018,
  author    = {Jacot, Arthur and Gabriel, Franck and Hongler, Cl{\'e}ment},
  title     = {Neural Tangent Kernel: Convergence and Generalization in Neural Networks},
  booktitle = {Advances in Neural Information Processing Systems},
  volume    = {31},
  publisher = {Curran Associates, Inc.},
  year      = {2018}
}

@inproceedings{Karimi2016,
  author    = {Karimi, Hamed and Nutini, Julie and Schmidt, Mark},
  title     = {Linear Convergence of Gradient and Proximal-Gradient Methods Under the Polyak-Lojasiewicz Condition},
  booktitle = {Joint European Conference on Machine Learning and Knowledge Discovery in Databases},
  series    = {Lecture Notes in Computer Science},
  volume    = {9851},
  pages     = {795--811},
  publisher = {Springer, Cham},
  year      = {2016}
}

@article{Polyak1963,
  author    = {Polyak, B. T.},
  title     = {Gradient methods for minimizing functionals},
  journal   = {Zhurnal Vychislitel'noi Matematiki i Matematicheskoi Fiziki},
  volume    = {3},
  number    = {4},
  pages     = {643--653},
  year      = {1963}
}

@article{Necoara2015,
  author    = {Necoara, Ion and Nesterov, Yurii and Glineur, Fran{\c{c}}ois},
  title     = {Linear convergence of first order methods for non-strongly convex optimization},
  journal   = {Mathematical Programming},
  volume    = {175},
  number    = {1},
  pages     = {69--107},
  year      = {2019}
}

@inproceedings{Arora2019a,
  author    = {Arora, Sanjeev and Du, Simon S. and Hu, Wei and Li, Zhiyuan and Salakhutdinov, Russlan and Wang, Ruosong},
  title     = {On Exact Computation with an Infinitely Wide Neural Net},
  booktitle = {Advances in Neural Information Processing Systems},
  volume    = {32},
  pages     = {8139--8148},
  year      = {2019}
}

@inproceedings{AllenZhu2019,
  author    = {Allen-Zhu, Zeyuan and Li, Yuanzhi and Song, Zhao},
  title     = {A Convergence Theory for Deep Learning via Over-Parameterization},
  booktitle = {Proceedings of the 36th International Conference on Machine Learning (ICML)},
  series    = {Proceedings of Machine Learning Research},
  volume    = {97},
  pages     = {242--252},
  publisher = {PMLR},
  year      = {2019}
}

@inproceedings{choromanska2015loss,
  author  = {Choromanska, Anna and Henaff, Mikael and Mathieu, Michael and Ben Arous, G{\'e}rard and LeCun, Yann},
  title   = {The Loss Surfaces of Multilayer Networks},
  booktitle = {Proceedings of the 18th International Conference on Artificial Intelligence and Statistics (AISTATS 2015)},
  pages   = {192--204},
  year    = {2015},
  address = {San Diego, CA}
}

@inproceedings{nguyen2017loss,
  author    = {Nguyen, Quynh and Hein, Matthias},
  title     = {The Loss Surface of Deep and Wide Neural Networks},
  booktitle = {Proceedings of the 34th International Conference on Machine Learning (ICML)},
  series    = {Proceedings of Machine Learning Research},
  volume    = {70},
  pages     = {2603--2612},
  publisher = {PMLR},
  year      = {2017}
}

@article{petzka2021nonattracting,
  author  = {Petzka, Henning and Sminchisescu, Cristian},
  title   = {Non-attracting Regions of Local Minima in Deep and Wide Neural Networks},
  journal = {Journal of Machine Learning Research},
  volume  = {22},
  number  = {143},
  pages   = {1--34},
  year    = {2021}
}

@inproceedings{kawaguchi2016deep,
  author    = {Kawaguchi, Kenji},
  title     = {Deep Learning without Poor Local Minima},
  booktitle = {Advances in Neural Information Processing Systems},
  volume    = {29},
  publisher = {Curran Associates, Inc.},
  year      = {2016}
}

@inproceedings{kawaguchi2020practical,
  author    = {Kawaguchi, Kenji and Huang, Jiaoyang},
  title     = {Gradient Descent Finds Global Minima for Generalizable Deep Neural Networks of Practical Sizes},
  booktitle = {57th Annual Allerton Conference on Communication, Control, and Computing (Allerton)},
  pages     = {92--99},
  publisher = {IEEE},
  year      = {2019}
}

@inproceedings{zhou2021local,
  author    = {Zhou, Mo and Ge, Rong and Jin, Chi},
  title     = {A Local Convergence Theory for Mildly Over-Parameterized Two-Layer Neural Network},
  booktitle = {Proceedings of the 34th Annual Conference on Learning Theory (COLT)},
  series    = {Proceedings of Machine Learning Research},
  volume    = {134},
  pages     = {4577--4632},
  publisher = {PMLR},
  year      = {2021}
}

@inproceedings{novak2022fast,
  author    = {Novak, Roman and Sohl-Dickstein, Jascha and Schoenholz, Samuel S.},
  title     = {Fast Finite Width Neural Tangent Kernel},
  booktitle = {Proceedings of the 39th International Conference on Machine Learning (ICML)},
  series    = {Proceedings of Machine Learning Research},
  volume    = {162},
  publisher = {PMLR},
  year      = {2022}
}

@article{xu2025local,
  author  = {Xu, Ziqing and Min, Hancheng and Tarmoun, Salma and Mallada, Enrique and Vidal, Rene},
  title   = {A Local Polyak-L{\'o}jasiewicz and Descent Lemma of Gradient Descent for Overparametrized Linear Models},
  journal = {Transactions on Machine Learning Research},
  year    = {2025}
}

@article{an2024convergence,
  author  = {An, Jing and Lu, Jianfeng},
  title   = {Convergence of Stochastic Gradient Descent under a Local Lojasiewicz Condition for Deep Neural Networks},
  journal = {Journal of Machine Learning},
  volume  = {4},
  number  = {2},
  pages   = {89--107},
  year    = {2025},
  doi     = {10.4208/jml.240724},
  note    = {Originally arXiv:2304.09221, 2023}
}

@inproceedings{wang2024monotone,
  author    = {Wang, Ruigang and Dvijotham, Krishnamurthy Dj and Manchester, Ian},
  title     = {Monotone, Bi-Lipschitz, and Polyak-{\L}ojasiewicz Networks},
  booktitle = {Proceedings of the 41st International Conference on Machine Learning},
  series    = {Proceedings of Machine Learning Research},
  volume    = {235},
  pages     = {50379--50399},
  year      = {2024},
  publisher = {PMLR}
}

@article{delarue2025genericity,
  author  = {Daudin, Samuel and Delarue, Fran{\c{c}}ois},
  title   = {Genericity of Polyak-Lojasiewicz Inequalities for Entropic Mean-Field Neural ODEs},
  journal = {arXiv preprint arXiv:2507.08486},
  year    = {2025}
}

@inproceedings{Du2019,
  title = {Gradient Descent Provably Optimizes Over-parameterized Neural Networks},
  author = {Du, Simon S. and Zhai, Xiyu and Poczos, Barnabas and Singh, Aarti},
  booktitle = {International Conference on Learning Representations (ICLR)},
  year = {2019}
}

@inproceedings{Du2018,
  author    = {Du, Simon S. and Lee, Jason D. and Li, Haochuan and Wang, Liwei and Zhai, Xiyu},
  title     = {Gradient Descent Finds Global Minima of Deep Neural Networks},
  booktitle = {Proceedings of the 36th International Conference on Machine Learning (ICML)},
  series    = {Proceedings of Machine Learning Research},
  volume    = {97},
  pages     = {1675--1685},
  publisher = {PMLR},
  year      = {2019}
}

@inproceedings{Lee2019,
  title = {Wide Neural Networks of Any Depth Evolve as Linear Models Under Gradient Descent},
  author = {Lee, Jaehoon and Xiao, Lechao and Schoenholz, Samuel S. and Bahri, Yasaman and Novak, Roman and Sohl-Dickstein, Jascha and Pennington, Jeffrey},
  booktitle = {Advances in Neural Information Processing Systems},
  volume = {32},
  year = {2019}
}

@article{Liu2020,
  title = {Loss Landscapes and Optimization in Over-Parameterized Non-Linear Systems and Neural Networks},
  author = {Liu, Chaoyue and Zhu, Libin and Belkin, Mikhail},
  journal = {Applied and Computational Harmonic Analysis},
  volume = {59},
  pages = {85--116},
  year = {2022},
  note = {Originally arXiv:2003.00307, 2020}
}

@article{Oymak2019,
  title   = {Toward Moderate Overparameterization: Global Convergence Guarantees for Training Shallow Neural Networks},
  author  = {Oymak, Samet and Soltanolkotabi, Mahdi},
  journal = {IEEE Journal on Selected Areas in Information Theory},
  volume  = {1},
  number  = {1},
  pages   = {84--105},
  year    = {2020},
  doi     = {10.1109/JSAIT.2020.2991332}
}

@inproceedings{he2016deep,
  author    = {He, Kaiming and Zhang, Xiangyu and Ren, Shaoqing and Sun, Jian},
  title     = {Deep Residual Learning for Image Recognition},
  booktitle = {Proceedings of the IEEE Conference on Computer Vision and Pattern Recognition (CVPR)},
  pages     = {770--778},
  year      = {2016}
}

@inproceedings{wu2018group,
  author    = {Wu, Yuxin and He, Kaiming},
  title     = {Group Normalization},
  booktitle = {Proceedings of the European Conference on Computer Vision (ECCV)},
  pages     = {3--19},
  year      = {2018}
}

@techreport{krizhevsky2009learning,
  author      = {Krizhevsky, Alex},
  title       = {Learning Multiple Layers of Features from Tiny Images},
  institution = {University of Toronto},
  year        = {2009}
}

@inproceedings{ioffe2015batch,
  author    = {Ioffe, Sergey and Szegedy, Christian},
  title     = {Batch Normalization: Accelerating Deep Network Training by Reducing Internal Covariate Shift},
  booktitle = {Proceedings of the 32nd International Conference on Machine Learning (ICML)},
  series    = {Proceedings of Machine Learning Research},
  volume    = {37},
  pages     = {448--456},
  publisher = {PMLR},
  year      = {2015}
}
\bibliographystyle{tmlr}

\end{document}